\documentclass{article}

\usepackage{microtype}
\usepackage{graphicx}
\usepackage{subfigure}
\usepackage{booktabs} 

\usepackage{hyperref}

\usepackage[accepted]{icml2025}

\usepackage{amsmath}
\usepackage{amssymb}
\usepackage{mathtools}
\usepackage{amsthm}
\usepackage{multirow}

\usepackage[capitalize,noabbrev]{cleveref}

\theoremstyle{plain}

\theoremstyle{definition}

\theoremstyle{remark}

\usepackage{colortbl}   
\usepackage{xcolor}

\usepackage[textsize=tiny]{todonotes}

\usepackage{tcolorbox}
\tcbuselibrary{most}

\newtcolorbox{examplebox}[1]{
  colback=white,
  colframe=gray!50!black,
  fonttitle=\bfseries,
  title=#1,
  width=\textwidth
}

\usepackage{tikz}

\icmltitlerunning{Think Twice, Act Once: A Co-Evolution Framework of LLM and RL for Large-Scale Decision Making}

\begin{document}

\twocolumn[
\icmltitle{Think Twice, Act Once: \\ A Co-Evolution Framework of LLM and RL for Large-Scale Decision Making}

\icmlsetsymbol{equal}{*}

\begin{icmlauthorlist}
\icmlauthor{Xu Wan}{1}
\icmlauthor{Wenyue Xu}{2}
\icmlauthor{Chao Yang}{3}
\icmlauthor{Mingyang Sun}{4}
\end{icmlauthorlist}

\icmlaffiliation{1}{Zhejiang University}
\icmlaffiliation{2}{Tongji University}
\icmlaffiliation{3}{Alibaba DAMO Academy}
\icmlaffiliation{4}{Peking University}

\icmlcorrespondingauthor{Mingyang Sun}{smy@pku.edu.cn}


\icmlkeywords{Machine Learning, ICML}

\vskip 0.3in
]



\printAffiliationsAndNotice{}  

\begin{abstract}
Recent advancements in Large Language Models (LLMs) and Reinforcement Learning (RL) have shown significant promise in decision-making tasks. Nevertheless, for large-scale industrial decision problems, both approaches face distinct challenges: LLMs lack real-time long-sequence decision-making capabilities, while RL struggles with sample efficiency in vast action spaces. To bridge this gap, we propose \textbf{A}gents \textbf{C}o-\textbf{E}volution (ACE), a synergistic framework between LLMs and RL agents for large-scale decision-making scenarios. 
ACE introduces a dual-role trajectory refinement mechanism where LLMs act as both \textit{Policy Actor} and \textit{Value Critic} during RL's training: the \textit{Actor} refines suboptimal actions via multi-step reasoning and environment validation, while the \textit{Critic} performs temporal credit assignment through trajectory-level reward shaping. Concurrently, RL agent enhances LLMs' task-specific decision-making with high-quality fine-tuning datasets generated via prioritized experience replay.
Through extensive experiments across multiple power grid operation challenges with action spaces exceeding 60K discrete actions, ACE demonstrates superior performance over existing RL methods and LLM-based methods.

\end{abstract}

\section{Introduction}
\label{Introduction}
Making effective control in large-scale physical systems has been a long-standing goal in artificial intelligence research \cite{stoica2017berkeley}. Such decision-making tasks, including traffic control \cite{zhang2022expression, du2023safelight}, power system operating \cite{yoon2021winning, dorfer2022power, chauhan2023powrl}, and multi-robot coordination \cite{mahler2019learning, kalashnikov2022scaling}, require sophisticated reasoning capabilities and rapid response mechanisms. Reinforcement Learning (RL) has been extensively studied as a promising approach to tackle these challenges for a long time. Through iterative interaction with the environment, RL agents learn optimal control policies by maximizing cumulative rewards. However, the paradigm of learning from scratch without prior knowledge \cite{sutton1999reinforcement} making RL agents inherently suffer from sample inefficiency. Moreover, the time-varying and stochastic characteristics of large-scale industrial scenarios result in a substantial gap between the converged solutions obtained through RL models and true optimality \cite{nian2020review}. 

Various approaches leverage expert knowledge to alleviate the inefficiency and sub-optimality intrinsic in RL training. In terms of policy guidance, a straightforward yet effective approach is Learning from Demonstration (LFD) \cite{argall2009survey}. LFD-based methods directly mimics expert demonstrations through behavior cloning or tries to optimize the reward function through Inverse Reinforcement Learning (IRL) \cite{ng2000algorithms, nair2018overcoming, torabi2018behavioral, adams2022survey}. While theoretically promising, LFD-based methods heavily rely on the quality of expert demonstrations and struggle to generalize to unseen scenarios, often suffering from distribution shift problems when encountering states not covered in the demonstration data.
Meanwhile, Human-in-the-Loop (HITL) approaches \cite{abel2017agent} has emerged as a promising paradigm that leverages expert feedback to adaptive guide policy learning. These methods incorporate human guidance either through real-time interaction \cite{knox2009interactively, macglashan2017interactive} or offline preference alignment \cite{bai2022training, dai2023safe}, enabling human trainers to provide evaluative feedback or corrective advice during policy learning. However, due to the unpredictable and subjective nature of human feedback, HITL-based approaches face key challenges including intensive human time requirements, feedback inconsistency, and the balance between human intervention and autonomous learning \cite{kumar2024applications}.

Recent advances in Large Language Models (LLMs) have opened new possibilities for enhancing RL frameworks. Through training on massive text data, LLMs have acquired rich world knowledge and reasoning capabilities, making them promising candidates for decision-making tasks \cite{lu2024chameleon, wang2023tpe}. Researchers have explored multiple LLM-RL integration paradigms: 1) semantic action space compression through LLM-guided abstraction \cite{zhou2023large}, 2) reward shaping via natural language instruction following \cite{tan2024true}, and 3) code generation for policy implementation \cite{du2023guiding}. Recent work like Thought Cloning \cite{liu2024rl} further demonstrates how LLMs can generate chain-of-thought trajectories to guide policy learning through structured reasoning traces. However, these approaches face significant challenges in industrial control scenarios.
While LLMs excel at high-level strategic planning, they struggle with long-sequence decision-making required for fine-grained control. Industrial tasks exhibit inherent time delays and temporal coupling, making it challenging for LLMs to generate coherent sequences of decisions over extended time horizons. Additionally, the autoregressive nature of transformer-based LLMs introduces significant latency \cite{fu2024serverlessllm}, making them impractical for real-time control loops requiring sub-1000ms responses. Our analysis suggests that direct LLM-based action generation fundamentally conflicts with the precision and timing requirements of industrial control systems, necessitating a new paradigm that combines LLMs' strategic reasoning with traditional RL's numerical optimization strengths.

\begin{figure*}[t]
\vskip 0.2in
\begin{center}
\centerline{\includegraphics[width=2\columnwidth]{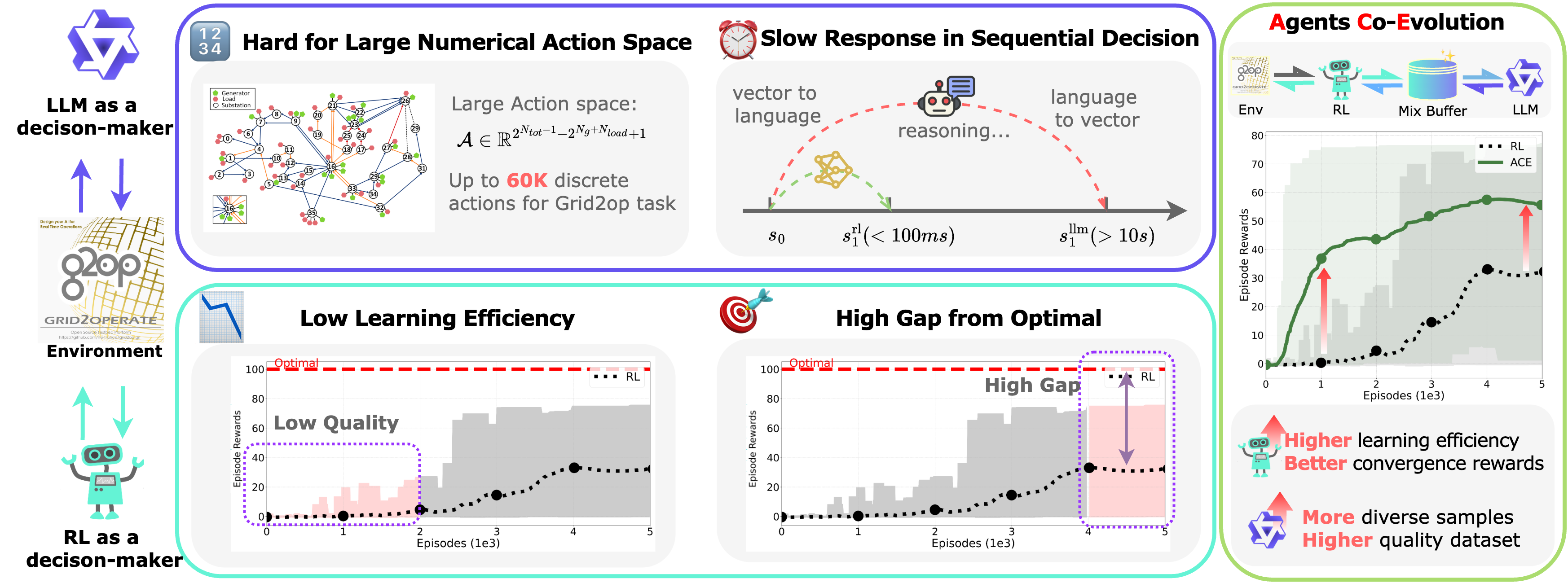}}
\caption{The motivation and challenges of integrating LLM with RL for large-scale industrial decision making.}
\label{fig:motivation}
\end{center}
\vskip -0.2in
\end{figure*}

Motivated by these observations, we develop a more suitable LLM-RL collaborative framework for industrial decision-making.
Unlike previous works that integrate LLMs during inference, we restrict LLM as an offline guider in the training phase, preserving real-time performance during online interaction and deployment.
Specifically, in the ``\textit{think twice}'' phase, LLMs guide RL policy updates through two key mechanisms: (1) rectifying interaction trajectories as a \textit{Policy Actor} to refine suboptimal decisions, and (2) performing trajectory-level reward shaping as a \textit{Value Critic} to enable better credit assignment. In the ``\textit{act once}'' phase, RL agent interacts with the environment to generate RL trajectories, which are then combined with LLM-refined trajectories to create a mixed experience buffer. This buffer serves both as a training buffer for RL and as a fine-tuning dataset to enhance LLM's task-specific guidance capabilities. To summarize, our main contributions are:

(1) We propose \textbf{A}gents \textbf{C}o-\textbf{E}volution (ACE), a co-evolution framework that separates LLM reasoning and RL execution into offline training and online deployment, enabling both effective learning and real-time decision-making in large-scale industrial scenarios.

(2) For guiding efficient exploration in large-scale decision-making, we develop a dual-role trajectory refinement mechanism where LLMs serve as both Policy Actor and Value Critic, addressing sample inefficiency and solution sub-optimality, respectively.

(3) For enabling continuous improvement of RL and LLMs, we establish an automated high-quality dataset generation workflow through reward-based prioritization and weighted policy update strategies in experience gathering.

(4) We demonstrate state-of-the-art performance on 3 L2RPN competitions, outperforming existing expert-guided RL methods and LLM methods in industrial environments with over 60K action space.

\section{Related Work}

\textbf{Expert-Guided Reinforcement Learning}
Expert guidance has been widely explored to address the sample efficiency and exploration challenges inherent in RL. Imitation learning-based methods, like DAgger \cite{ross2011reduction}, Hg-DAgger \cite{le2018hierarchical}, Soft DAgger \cite{nazeer2023soft} iteratively collect expert feedback on states visited by the learning policy to address distribution shift. Instead of direct action imitation, IRL-based methods select a reward function from the set of possible solutions that best explains expert behavior. AIRL \cite{fu2017learning}, MaxEnt-IRL \cite{zeng2022maximum}, and Offline IRL \cite{zeng2023understanding} leveraging different optimization principles to recover more robust reward functions. For industrial scenarios with large action spaces, \cite{yoon2021winning} reduces the action space based on sample frequency of offline expert data, while \cite{chauhan2023powrl} employs predefined rules to simplify RL decision-making tasks. However, these methods often sacrifice the exploration of optimal solutions. Alternatively, \cite{dorfer2022power} utilizes monte-carlo tree search to guide policy exploration through look-ahead planning, but it incurs significant computational overhead from extensive simulations.

Unlike previous approaches that rely on extensive offline expert data or look-ahead simulation, the ACE framework leverages LLM's superior reasoning and in-context learning capabilities to perform trajectory refinement and reward shaping from RL's demonstrations.

\textbf{Language Models in Decision-Making}
Focusing on combining LLMs' capabilities in control tasks, several works adapt LLMs to directly generate executable actions. For instance, CALM \cite{yao2020keep}, TWOSOME \cite{tan2024true} and POAD \cite{wen2024reinforcing} constrain the action space to a restricted subset using LLMs and employ RL to align agents' knowledge with specific environments. Similarly, various approaches decompose complex tasks into manageable subgoals. SayCan \cite{ahn2022can}, LgTs \cite{shukla2023lgts} and DART-LLM \cite{wang2024dart} break down high-level instructions into executable skills using affordance functions and validity scores, while ReAct \cite{yao2022react} utilizes chain-of-thought prompting to generate task-specific actions with reasoning traces.

Furthermore, recent studies have explored integrating LLMs as policy experts to guide RL agents' interactions \cite{nam2023lift}. LLM4Teach \cite{zhou2023large} incorporates LLM guidance by introducing policy regularization terms in the RL optimization process. ELLM \cite{du2023guiding} guides RL policy pre-training through LLM-suggested goals, while LINVIT \cite{zhang2024can} incorporates LLM guidance as a regularization factor for value-based RL. Most closely aligned with our motivation is RL-GPT \cite{liu2024rl}, which unifies coding and learning optimization in the RL training pipeline to help RL systems learn better decision-making strategies. However, RL-GPT relies on continuous LLM interaction during game tasks, resulting in training costs and real-time requirements that exceed the constraints of industrial applications.

\section{Method}
In this section, we describe the ACE framework, consisting of 1) direct policy learning through RL's environmental interaction, 2) trajectory refinement through LLM's dual roles as Actor and Critic, and 3) experience gathering that enables effective co-evolution between the two agents. The pseudo-code of ACE is shown in Algorithm \ref{alg::ace}. 
\subsection{First Think: Direct Policy Learning through Environment Interaction}

We first formulate the sequential decision-making problem as a Markov Decision Process (MDP) \cite{puterman1990markov} defined by tuple $(\mathcal{S}, \mathcal{A}, P, R, \gamma)$, where $\mathcal{S}$ and $\mathcal{A}$ denote the state and action spaces, respectively. $P: \mathcal{S} \times \mathcal{A} \times \mathcal{S} \rightarrow [0,1]$ is the transition probability, $R: \mathcal{S} \times \mathcal{A} \rightarrow \mathbb{R}$ is the reward function, and $\gamma \in (0,1]$ is the discount factor.

\paragraph{Optimization Objective.} Efficient policy exploration and experience collection are crucial for handling large-scale numerical action spaces. Therefore, we adopt the Soft Actor-Critic (SAC) \cite{haarnoja2018soft} algorithm as our RL module through off-policy training for better experience utilization. The SAC agent aims to learn a policy $\pi$ that maximizes both the expected return and policy entropy. Our objective function is:
\begin{equation}
    J(\pi) = \mathbb{E}_{\tau\sim\pi}\left[\sum_{t=0}^{\infty} \gamma^t (R(s_t,a_t) + \alpha\mathcal{H}(\pi(\cdot|s_t)))\right]
\end{equation}
where $\mathcal{H}(\pi(\cdot|s_t)) = -\sum_a \pi(a|s_t)\log \pi(a|s_t)$ represents the entropy of the policy $\pi$ at state $s_t$, $\alpha$ is the temperature parameter controlling the trade-off between exploration and exploitation.

\paragraph{Learning from Interaction.} During standard environment interaction, transition tuples $\tau = (s_t, a_t, r_t, s_{t+1}, d)$ are stored into a replay buffer $\mathcal{D}_{\text{RL}}$ for each timestep. Each tuple consists of the current state $s_t$, the action $a_t$ sampled from RL's policy $\pi(a_t|s_t)$, the received reward $r_t$, and the next state $s_{t+1}$. To learn from experiences, a mini-batch of transitions is sampled from $\mathcal{D}_{\text{RL}}$ to optimize both the Q-function $Q_\phi$ and policy $\pi_\theta$. Specifically, the Q-function is updated by minimizing:
\begin{equation}
\begin{aligned}
   \mathcal{L}_Q(\phi) &= \mathbb{E}_{(s,a,r,s')\sim\mathcal{D}_{\text{RL}}}\left[(Q_\phi(s,a) - y)^2\right] \\
   y &= r + \gamma(Q_{\phi'}(s',a') - \alpha\log \pi_\theta(a'|s'))
\end{aligned}
\label{eq: SAC_L_Q}
\end{equation}
where $\phi'$ denotes the target network parameters and $a'$ is sampled from the current policy $\pi_\theta$. The policy is then improved by minimizing:
\begin{equation}
   \mathcal{L}_\pi(\theta) = \mathbb{E}_{s\sim\mathcal{D}_{\text{RL}}}\left[\mathbb{E}_{a\sim\pi_\theta}[\alpha \log \pi_\theta(a|s) - Q_\phi(s,a)]\right]
    \label{eq: SAC_L_a}
\end{equation}

As shown in Eq. (\ref{eq: SAC_L_Q}) and Eq. (\ref{eq: SAC_L_a}), the learning efficiency heavily depends on the target value $y$ computed using both Q-function and policy. However, in early training stages with large action spaces, \textit{both the Q-function estimation and policy exploration are unreliable.} It directly motivates our second think mechanism, which leverages LLM's reasoning capabilities to enhance the quality of experiences in $\mathcal{D}_{\text{RL}}$.

\begin{figure}[t]
\vskip 0.2in
\begin{center}
\centerline{\includegraphics[width=\columnwidth]{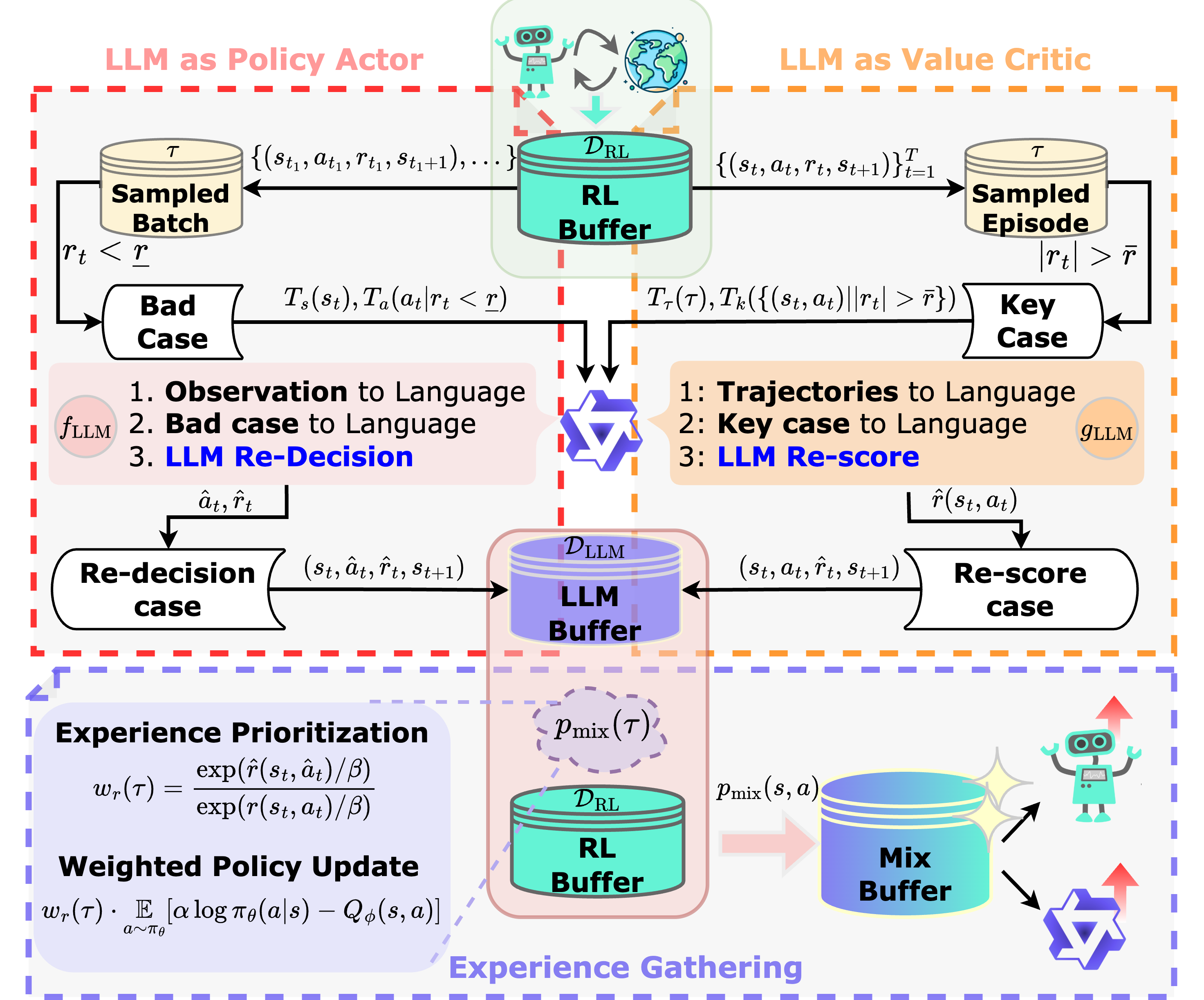}}
\caption{The architecture of ACE framework with dual-role LLM.}
\label{fig:Framework}
\end{center}
\vskip -0.2in
\end{figure}

\subsection{Second Think: Trajectory Refinement through LLMs}

In this section, we introduce a dual-role refinement framework in which LLMs function as both an Actor for policy refinement and a Critic for value re-assessment, as shown in Figure~\ref{fig:Framework}.
It is driven by two key insights: (1) LLMs' strong in-context reasoning capabilities in inferring better alternative policy from failure cases; (2) LLMs' counterfactual reasoning abilities facilitate better credit assignment over long-term dependencies.

\paragraph{LLMs as Policy Actor} When activated, LLMs serve as a policy actor to refine suboptimal decisions. For mini-batches sampled from the replay buffer $\mathcal{D}_{\text{RL}}$, we identify transitions where the reward $r < \underline{r}$, marking them as inappropriate decisions that require refinement. These states are converted into natural language descriptions and provided to the LLM along with essential context:
\begin{equation}
\hat{a}_t = f_{\text{LLM}}(P_r, T_s(s_t), T_a(a_t)|r(s_t, a_t) < \underline{r})
\label{eq::f_llm}
\end{equation}
where $P_r$ is the task description prefix that supplies necessary domain knowledge, $T_s(\cdot)$ and $T_a(\cdot)$ denote the text conversion functions that transform states and actions into natural language descriptions, respectively. The refined action $\hat{a}_t$ is then selected as the action from the LLM's output and subsequently validated through environment simulation to obtain new transitions $(s_t, \hat{a}_t, r_t(s_t, \hat{a}_t), \hat{s}_{t+1}, \hat{d})$, which are stored in a separate LLM buffer $\mathcal{D}_{\text{LLM}}$.

\paragraph{LLMs as Value Critic} In later training stages, we leverage LLMs to perform trajectory-level counterfactual reasoning and re-evaluate the long-term impact of key decisions:
\begin{equation}
\hat{r}_t = g_{\text{LLM}}(P_r, T_{\tau}(\tau), T_k(\{(s_t, a_t)||r(s_t, a_t)| > \bar{r}\}))
\label{eq::g_llm}
\end{equation}
where $\tau = \{(s_t, a_t, r_t, s_{t+1},d)\}_{t=1}^T$ represents the complete episode trajectory, $\bar{r}$ denotes the threshold for identifying potentially critical trajectories.
This mechanism can be viewed as an extension of implicit multi-step TD($\lambda$) \cite{sutton2018reinforcement}: LLMs approximate eligibility traces through trajectory-level reasoning, enabling credit assignment over broader temporal horizons. Compared to TD($\lambda$)'s exponential decay assumption, LLM's causal attribution provides non-parametric modeling capability for long-term dependencies.

To maintain stability while allowing meaningful adjustments, we discretize the reward modifications into four levels: $\{-2K, -K, +K, +2K\}$ and limit the number of modifications per episode, where $K$ is a predefined adjustment scale.

\subsection{Co-evolution through Experience Gathering}
To enable effective co-evolution between RL and LLM, we further construct a mixed buffer $\mathcal{D}_{\text{mix}}$ that serves both RL policy training and LLM online fine-tuning through the following sampling strategy.

Let $\beta$ denote the ratio of samples drawn from the LLM buffer and $\tau$ denote the sampled transitions. The base mixing distribution is:
\begin{equation}
p_{\text{mix}}(\tau) = (1-\beta)p_{\text{RL}}(\tau) + \beta p_{\text{LLM}}(\tau)
\end{equation}
where \( p_{\text{RL}}(\tau) \) and \( p_{\text{LLM}}(\tau) \) represent the sampling distributions for $\mathcal{D}_{\text{RL}}$ and $\mathcal{D}_{\text{LLM}}$, respectively.

To ensure the quality of LLM-refined experiences, we employ two key mechanisms:

\paragraph{Reward-based Experience Prioritization} First, we evaluate the quality of LLM refinements based on the immediate outcomes and termination signal:
\begin{equation}
\mathbb{I}v(\tau) = \mathbb{I}[\hat{r}(s,\hat{a}) \geq 0 \land \neg d(\hat{s'})]
\end{equation}
where $d(\hat{s'}) \in \{0,1\}$ indicates whether the next state $\hat{s'}$ after LLM refine leads to episode termination (1) or continuation (0), and $\hat{r}(s,\hat{a}) \geq 0$ refers to the non-negative reward after LLM's refinement.

To prioritize high-reward experiences during sampling, we introduce a reward-based importance weight:
\begin{equation}
w_r(\tau) = \frac{\exp(\hat{r}(s,\hat{a})/\beta)}{\exp(r(s,a)/\beta)}
\label{eq::w_r}
\end{equation}
where $w_r(\tau)$ serves as a prior for experience replay, ensuring that the refined transitions by $f_{\text{LLM}}$ or $g_{\text{LLM}}$ leading to higher rewards than RL agent are more likely to be sampled.

Therefore, the complete mixed sampling distribution is re-formulated as:
\begin{equation}
p_{\text{mix}}(\tau) =
\begin{cases}
\frac{1-\beta}{|\mathcal{D}_{\text{RL}}|} &\text{if } \tau \in  \mathcal{D}_{\text{RL}} \\
\frac{\beta \cdot \mathbb{I}v(\tau) \cdot w_r(\tau)}{\sum_{\substack{\tau \in \mathcal{D}_{\text{LLM}}}} \mathbb{I}v(\tau) \cdot w_r(\tau)} & \text{if } \tau \in \mathcal{D}_{\text{LLM}}
\end{cases}
\label{eq::mix_buffer}
\end{equation}
\paragraph{Reward-weighted Policy Learning} To prioritize learning from valuable LLM refinements, we introduce a reward-based weighting mechanism for RL agent:

\begin{equation}
    \begin{aligned}
\mathcal{L}_{\pi}(\theta) = \underset{\tau \sim p_{\text{mix}}}{\mathbb{E}}[w_r(\tau) \cdot \underset{a\sim\pi_\theta}{\mathbb{E}}[\alpha \log \pi_\theta(a|s) - Q_\phi(s,a)]]
\end{aligned}
\label{eq::rl_loss}
\end{equation}
$w_r(\tau)$ naturally emphasizes LLM refinements that lead to higher rewards, allowing the RL agent to focus on learning from more valuable demonstrations.

For the LLM module, we apply low-rank adaptation fine-tuning on $\mathcal{D}_{\text{mix}}$ with reward signals as labels to enhance the task-specific capabilities.

\begin{algorithm}[t]
\caption{\textbf{A}gents \textbf{C}o-\textbf{E}volution (ACE) Framework}
\label{alg::ace}
\begin{algorithmic}[1]
\REQUIRE Initial RL policy $\pi_\theta$, Q-function $Q_\phi$, model $f_{\text{LLM}}$ and $g_{\text{LLM}}$, buffers $\mathcal{D}_{\text{RL}}$, $\mathcal{D}_{\text{LLM}}$
\STATE // \textit{Act Once}
\FOR{each environment step}
   \STATE Sample action $a_t \sim \pi_\theta(\cdot|s_t)$
   \STATE Execute $a_t$, observe $r_t, s_{t+1}$, done marker $d_t$
   \STATE Store $\tau:=(s_t, a_t, r_t, s_{t+1}, d_t)$ in $\mathcal{D}_{\text{RL}}$
\ENDFOR
\STATE // \textit{Think Twice}

\FOR{each LLM active step}
    \STATE Sample batch $\tau \sim \mathcal{D}_{\text{RL}}$
   \IF{$f_{\text{LLM}}$ is active and $r_t < \underline{r}$ in $\tau$}
       \STATE Convert state-action to text: $T_s(s_t), T_a(a_t)$
       \STATE Get refined action $\hat{a}_t$ by Eq. (\ref{eq::f_llm})
   \ENDIF
   \IF{$g_{\text{LLM}}$ is active and $|r_t| > \bar{r}$ in $\tau$}
       \STATE Convert trajectories to text: $T_\tau(\tau)$
       \STATE Get refined rewards $\hat{r}_t$ by Eq. (\ref{eq::g_llm})
   \ENDIF
   \STATE Simulate by Grid2Op and store refined transitions $(s_t, \hat{a}_t, \hat{r}_t, \hat{s}_{t+1}, \hat{d_t})$ in $\mathcal{D}_{\text{LLM}}$
\ENDFOR

\FOR{each RL update step}
   \STATE Sample mixed batch according to $p_{\text{mix}}$ in Eq. (\ref{eq::mix_buffer})
   \STATE Compute importance weights $w_r(\tau)$ by Eq. (\ref{eq::w_r})
   \STATE Update Q-function by Eq.(\ref{eq: SAC_L_Q})
   \STATE Update RL policy with weighted loss by Eq. (\ref{eq::rl_loss})
\ENDFOR

\FOR{each LLM fine-tuning step}
   \STATE Generate buffer $\mathcal{D}_{\text{mix}}$ according to $p_{\text{mix}}$ in Eq. (\ref{eq::mix_buffer})
   \STATE Fine-tuning $f_{\text{LLM}}$ and $g_{\text{LLM}}$ using $\mathcal{D}_{\text{mix}}$
\ENDFOR
\end{algorithmic}
\end{algorithm}

\section{Experiments}

\subsection{Environmental Setup}

We evaluate ACE in three complex real-world power system operation cases. The data is sourced from the Grid2Op open-source platform \cite{donnot2020grid2op} provided by RTE France, Europe's largest grid operator. Specifically, we use datasets as follows:

\textbf{(1) L2RPN WCCI 2020 Challenge}  \cite{marot2020learning} This challenge presents a medium-sized power grid challenge that simulates one-third of the US Midwest grid. The environment consists of 36 substations, 59 lines, and 22 generators, with an extensive topology assignment action space \textit{exceeding 60,000 possible actions} \cite{yoon2021winning}.
The objective is twofold: (1) developing strategies to overcome operational obstacles like grid congestion, and (2) optimizing various operational costs including power line losses, dispatch costs, and outage costs. The challenge incorporates realistic scenarios such as load fluctuations and line maintenance. For evaluation, the test dataset comprises 10 different episodic scenarios, each spanning 3 days (864 steps) with varying difficulty levels.

\textbf{(2) L2RPN NeurIPS 2020 Challenge} \cite{marot2021learning}: This challenge extends the WCCI 2020 grid environment by introducing an adversarial setting. The key innovation is the addition of an ``opponent'' that follows a heuristic strategy of randomly disconnecting heavily loaded power lines.
The test dataset is conducted on 24 weekly episodes (2016 steps), with control decisions made at 5-minute intervals.

\textbf{(3) L2RPN WCCI 2022 Challenge} \cite{marot2022learning}: This competition utilizes the industry standard synthetic IEEE-118 power grid environment, featuring 118 substations, 186 power lines, 91 loads, and 62 generators. It introduces more renewable generators and storage systems than previous challenges, focusing on electricity production uncertainty and AI agent robustness. The action space is significantly larger, \textit{with over 70,000 discrete actions} for substation switching alone \cite{dorfer2022power}. The training dataset spans 32 years of grid data at 5-minute intervals (2016 steps), totaling approximately 1.7 GB.
    
\begin{figure*}[t]
\vskip 0.2in
\begin{center}
\centerline{\includegraphics[width=2 \columnwidth]{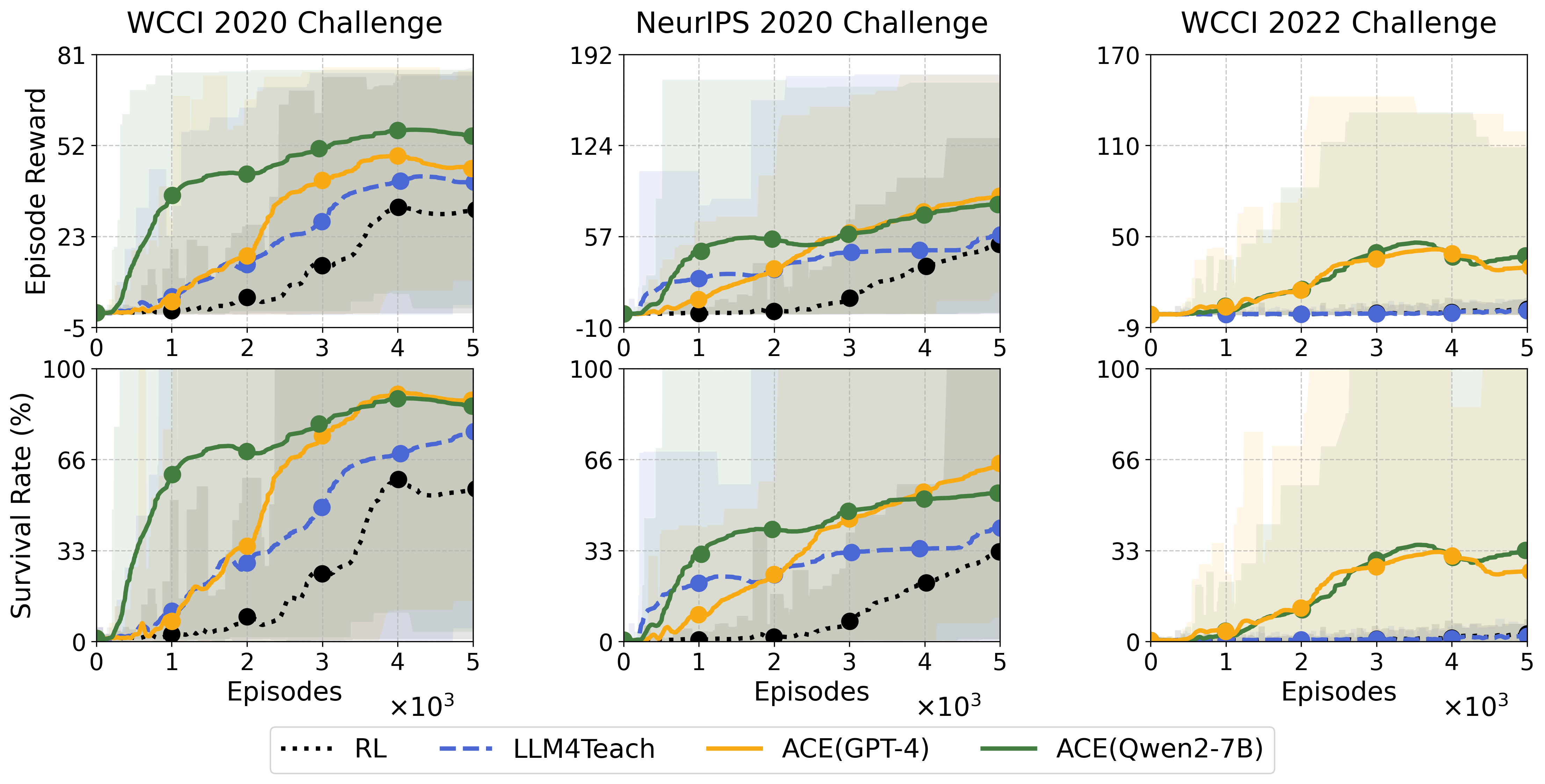}}
\caption{Training episode rewards comparison between ACE and baselines across 3 L2RPN competitions.}
\label{fig:Result}
\end{center}
\vskip -0.2in
\end{figure*}

\begin{table*}[ht]
\centering
\caption{Test Performance comparison across 3 L2RPN competitions. Results show mean $\pm$ standard deviation over 5 runs. Best results are in \textbf{bold}.}
\label{tab:performance}
\begin{tabular}{l l l c c c c}
\toprule
\textbf{Dataset} & \textbf{Method} & \textbf{Type} & \textbf{Episode} & \textbf{Survival} & \textbf{Sample} & \textbf{Test} \\
&  &  & \textbf{Rewards} & \textbf{Rate (\%)} & \textbf{Requirements} & \textbf{Time(s)} \\
\midrule
\multirow{5}{*}{\textbf{WCCI 2020}} & Expert-guided RL & RL & 57.1 $\pm$ 3.9 & 78.3 $\pm$ 3.9 & 100 K & 46.1 \\
& Qwen2-7B & LLM & 21.9 $\pm$ 1.9 & 30.5 $\pm$ 0.8 & N/A & 1480.8 \\
& GPT-4 & LLM & 29.5 $\pm$ 2.1 & 41.7 $\pm$ 1.5 & N/A & 3415.3 \\
& LLM4Teach & LLM+RL & 64.4 $\pm$ 4.2 & 78.5 $\pm$ 2.4 & 100 K & 46.2 \\
& ACE (Qwen2-7B + SFT)$^\dagger$ & LLM+RL & \cellcolor{gray!15} \textbf{69.8 $\pm$ 1.9} & \cellcolor{gray!15} \textbf{92.9 $\pm$ 0.0} & \cellcolor{gray!15} \textbf{40 K + 287} & \cellcolor{gray!15} 54.8 \\
& ACE (GPT-4) $^\dagger$ & LLM+RL & \cellcolor{gray!15} 67.7 $\pm$ 1.8 & \cellcolor{gray!15} 92.7 $\pm$ 0.0 & \cellcolor{gray!15} \textbf{40 K + 364} & \cellcolor{gray!15} \textbf{38.7} \\
\midrule  
\multirow{5}{*}{\textbf{NeurIPS 2020}} & Expert-guided RL & RL & 128.5 $\pm$ 5.1 & 72.9 $\pm$ 5.5 & 100 K & \textbf{217.2} \\
& Qwen2-7B & LLM & 37.5 $\pm$ 1.8 & 21.8 $\pm$ 1.0 & N/A & 10729.7 \\
& GPT-4 & LLM & 37.8 $\pm$ 1.9 & 42.3 $\pm$ 1.3 & N/A & 12949.8 \\
& LLM4Teach & LLM+RL & 131.5 $\pm$ 0.3 & 74.8 $\pm$ 0.2 & 100 K  & 212.9 \\
& ACE (Qwen2-7B + SFT) & LLM+RL & \cellcolor{gray!15} \textbf{145.3 $\pm$ 6.4} & \cellcolor{gray!15} \textbf{84.8 $\pm$ 3.7} & \cellcolor{gray!15} \textbf{40 K + 400} & \cellcolor{gray!15} 218.5 \\
& ACE (GPT-4) $^\dagger$ & LLM+RL & \cellcolor{gray!15} 143.5 $\pm$ 4.3 & \cellcolor{gray!15} 84.1 $\pm$ 4.9 & \cellcolor{gray!15} \textbf{40 K + 307} & \cellcolor{gray!15} 219.1 \\
\midrule
\multirow{5}{*}{\textbf{WCCI 2022}} & Expert-guided RL & RL & 31.0 $\pm$ 2.7 & 33.7 $\pm$ 2.0 & 200 K & \textbf{1440.0} \\
& Qwen2-7B & LLM & 16.7 $\pm$ 0.9 & 19.6 $\pm$ 2.6 & N/A & 22436.0 \\
& GPT-4 & LLM & 18.3 $\pm$ 0.8 & 22.2 $\pm$ 1.4 & N/A & 22188.6 \\
& LLM4Teach & LLM+RL & 23.6 $\pm$ 3.4 & 29.5 $\pm$ 2.0 & 200 K & 1116.8 \\
& ACE (Qwen2-7B + SFT) & LLM+RL & \cellcolor{gray!15} \textbf{75.9 $\pm$ 2.7} & \cellcolor{gray!15} \textbf{54.3 $\pm$ 2.0} & \cellcolor{gray!15} \textbf{50 K + 564} & \cellcolor{gray!15} 3627.2 \\
& ACE (GPT-4) $^\dagger$ & LLM+RL & \cellcolor{gray!15} 75.4 $\pm$ 6.3 & \cellcolor{gray!15} 49.5 $\pm$ 4.0 & \cellcolor{gray!15}\textbf{ 50 K + 682 }& \cellcolor{gray!15} 7478.9 \\
\bottomrule
\multicolumn{7}{p{1\textwidth}}{$^\dagger$ ``SFT'' refers to the process of fine-tuning LLM components using LoRA parameters with the mixed buffer $\mathcal{D}_{\text{mix}}$. ACE (GPT-4) does not include SFT, only using GPT for enhancing RL with our ACE framework.}
\end{tabular}
\end{table*}

\subsection{Baseline Methods}
We benchmark the ACE framework against state-of-the-art approaches across three categories:

\textbf{(1) Expert-guided RL}: We compare with the winning solutions from previous challenges. Building upon the WCCI 2020 winning solution \cite{yoon2021winning}, which leverages hierarchical policy and after-state representation, we adopt its architecture as our backbone but enhance it with our dual-role LLM refinement mechanism. 
   
\textbf{(2) LLM only}: We evaluate the direct decision-making performance of different pre-trained LLM models: GPT-4o-0806 \cite{achiam2023gpt} and Qwen2-7B-Instruct \cite{bai2023qwen} using the same prompts as ACE on the test set.

\textbf{(3) LLM-guided RL}: Given that existing LLM-RL methods typically operate during inference, which is impractical for L2RPN's large-scale tasks, we implement a modified version of LLM4Teach \cite{zhou2023large}: using KL-divergence constraints to regularize the RL policy $\pi_\theta$'s updates by minimizing the deviation from the LLM guider's policy distribution $\pi_{\text{LLM}}$.
    \begin{equation}
         \mathcal{L}_{\pi}(\theta) = \mathcal{L}_{\text{RL}}(\theta) + \lambda\mathbb{E}_{s\sim \tau} \text{KL}(\pi_{\text{LLM}}(\cdot|s)||\pi_\theta(\cdot|s))
         \label{eq::llm4teach}
    \end{equation}
    
As illustrated in Eq. (\ref{eq::llm4teach}), the modified LLM4Teach adapts the mode similar to ACE, where alignment is only enabled during the training phase with Qwen2-7B-Instruct as the aligned model.    

\subsection{Implementation Details}
We utilize Qwen2-7B instruct and GPT-4o-0806 as the ACE framework's LLM component, named ACE (Qwen2-7B) and ACE (GPT-4), respectively. 

For trajectory refinement, we introduce two effective tricks: bad case reasoning and multi-round reasoning mechanism. The \textit{bad case reasoning} mechanism integrates RL's bad actions of the current state into \(f_{\text{LLM}}\)'s prompt and formulates instructions that prompt LLMs to reason and avoid making similar suboptimal decisions. Moreover, \textit{multi-round reasoning} leverages the simulation functionality provided by Grid2Op to estimate the rewards following LLMs' refined actions. If the estimated rewards are inferior to RL agent, the system initiates multi-round decision-making, with a maximum of five rounds set in our experiments.

For efficient memory management, we limit the LLM buffer's maximum size to 256. When updating policy, we sample batches $\tau$ according to importance weights $w_r(\tau)$ from $\mathcal{D}_{\text{mix}}$ for RL training. In ACE (Qwen2-7B), we perform online fine-tuning of the LLM model for every 100 generated samples, while ACE (GPT-4) operates without fine-tuning. Besides, we activate the LLM Value Critic module $g_{\text{LLM}}$ only in the later stages of training  ($\mathcal{D}_{\text{LLM}}$ is full), where the modified rewards directly replace the original rewards without additional storage.

\subsection{Main Results}
We evaluate ACE against baseline methods across three L2RPN competition environments. As shown in Figure~\ref{fig:Result} and Table~\ref{tab:performance}, ACE consistently outperforms all baselines across different metrics.

From the perspective of \textbf{episode rewards}, \textit{ACE demonstrates substantial improvements across all three challenges, including the champion solutions from WCCI 2020 and Neurips 2020}. In the WCCI 2020 challenge, ACE achieves the highest episode reward, surpassing the pure RL approach by 22.2\% and pure LLM approaches by over 130\%. Similar results are observed in the NeurIPS 2020 challenge, where ACE achieves 145.3, outperforming other methods by at least 10.5\%. Most notably, in the more complex WCCI 2022 environment, ACE achieves a remarkable 145\% improvement over the Expert-guided RL baseline.

From the \textbf{decision-making efficiency} perspective, \textit{ACE maintains competitive real-time performance}. In the WCCI 2020 challenge, ACE (GPT-4) achieves a test time of 38.7s, which is comparable to the expert-guided RL baseline and significantly faster than pure LLM approaches. This phenomenon further demonstrates that for large-scale sequential decision-making problems with long time series, the running time of pure LLMs does not meet industrial requirements.

In terms of \textbf{sample efficiency}, we find that by \textit{injecting refinements for less than 700 selected samples, the convergence speed of RL can be significantly improved}. For the WCCI 2020 challenge, ACE requires only 287 LLM refinements to achieve state-of-the-art performance, compared to 100K samples needed by traditional approaches. This efficiency is also evident in the WCCI 2022 challenge, where ACE achieves superior performance with just 50K samples plus refinements, while baseline methods require 200K samples yet achieve lower performance.

\subsection{Ablation Studies}
To comprehensively assess the efficacy of each component within the ACE framework, we conduct a series of ablation studies of two pivotal modules, \(f_{\text{LLM}}\) and \(g_{\text{LLM}}\), as well as two reasoning strategies used for \(f_{\text{LLM}}\) interaction: multi-round reasoning and bad-case reasoning. The detailed results are summarized in Table \ref{tab:ablation} and Table \ref{tab:llm_ablation}.

\begin{table}[tb]
\centering
\small
\caption{Ablation study of ACE components on WCCI 2020 dataset.}
\label{tab:ablation}
\begin{tabular}{l c c}
\toprule
\textbf{ACE Variant} & \textbf{Episode Rewards} & \textbf{Survival Rate (\%)} \\
\midrule
Full ACE & \textbf{69.8 } & \textbf{92.9 } \\
\midrule
w/o $f_{\text{LLM}}$ & 48.3  & 71.4  \\
w/o $g_{\text{LLM}}$ & 61.5  & 84.7  \\
w/o Bad Case & 60.2  & 76.3  \\
w/o Multi-round & 65.7  & 88.5 \\
\bottomrule
\multicolumn{3}{p{0.45\textwidth}}{\small \textit{Note:} We use ACE as the baseline and limit w/o Multi-round Reasoning to process the same samples of RL trajectories.} 
\end{tabular}
\end{table}

First, we examine the impact of removing the core modules. As demonstrated in Table \ref{tab:ablation}, eliminating the Actor component \(f_{\text{LLM}}\) leads to the most pronounced performance decline, with the episode rewards plummeting to 48.3 and the survival rate dropping to 71.4\%. Similarly, the removal of the Critic component \(g_{\text{LLM}}\) also results in a notable decrease in performance, with episode rewards reaching 61.5 and a survival rate of 84.7\%.

\begin{figure}[t]
\begin{center}
\centerline{\includegraphics[width=\columnwidth]{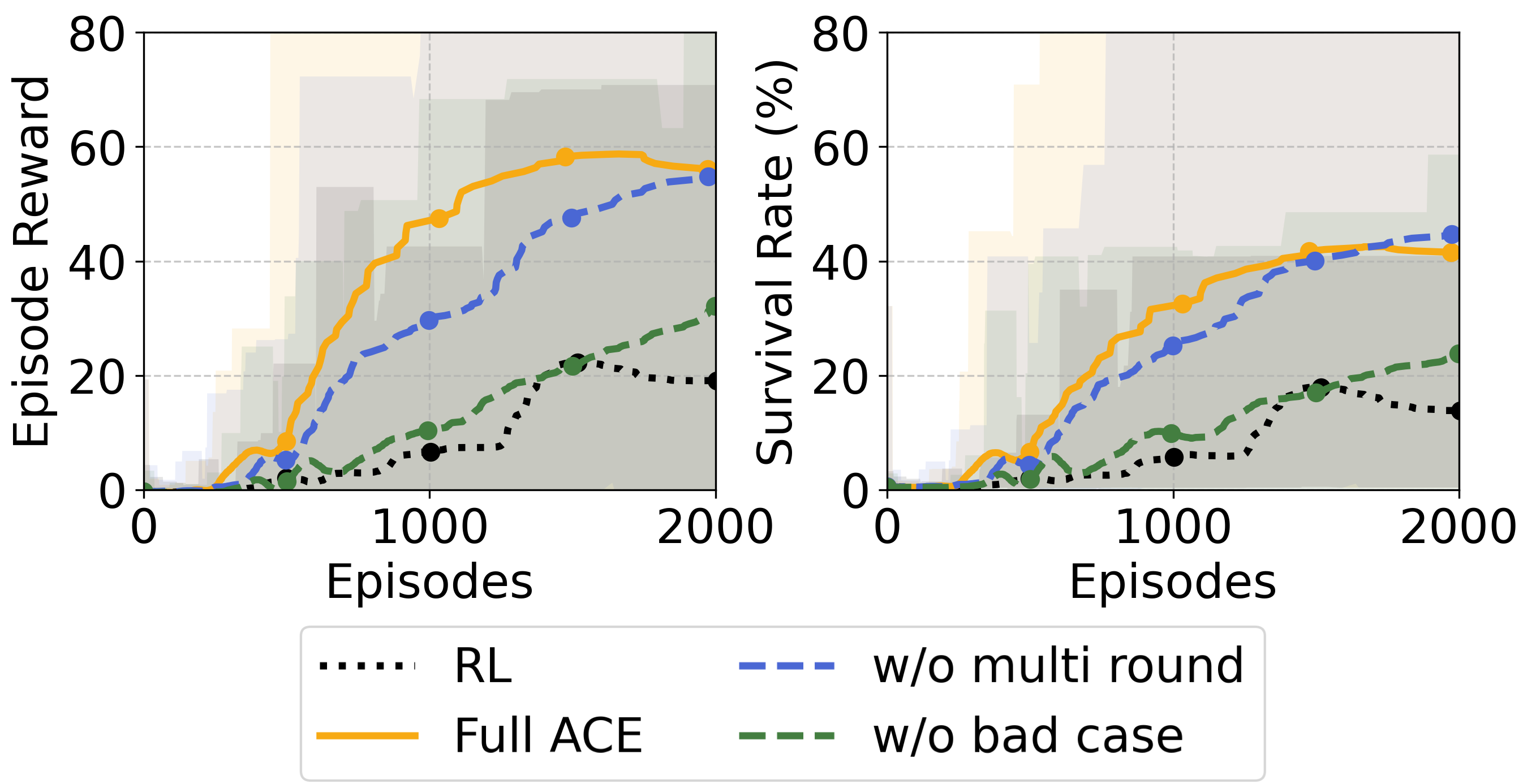}}
\caption{Ablation study results comparing ACE variants over training episodes on NeurIPS 2020 dataset.}
\label{fig:Ablation}
\end{center}
\vskip -0.2in
\end{figure}

Additionally, we analyze the impact of the reasoning strategies. The results in Figure. \ref{fig:Ablation} show that removing the bad case reasoning strategy reduces the rewards to 60.2 and the survival rate to 76.3\%, while the removal of multi-round reasoning leads to rewards of 65.7 and a survival rate of 88.5\%.
This demonstrates that using bad actions as negative examples effectively leverages the LLM's reasoning capabilities to generate superior decisions.

\begin{table}[tb]
\centering
\footnotesize
\caption{Ablation study of $f_{\text{LLM}}$ and $g_{\text{LLM}}$ Variants on WCCI 2020 dataset}
\label{tab:llm_ablation}
\resizebox{\linewidth}{!}{
\begin{tabular}{l c c l c c}
\toprule
\multicolumn{3}{c}{\textbf{$f_{\text{LLM}}$ Variants}} & \multicolumn{3}{c}{\textbf{$g_{\text{LLM}}$ Variants}} \\
\cmidrule(lr){1-3} \cmidrule(lr){4-6}
\textbf{Params} & \textbf{Reward} & \textbf{Survival} & \textbf{Params} & \textbf{Reward} & \textbf{Survival} \\
\midrule
\multicolumn{3}{c}{\textit{Active Frequency}} & \multicolumn{3}{c}{\textit{Active Frequency}} \\
128 & 66.2 & 89.1 & 32 & \textbf{68.1} & \textbf{94.0} \\
256 & \textbf{68.1} & \textbf{94.0} & 128 & 64.3 & 87.4 \\
512 & 63.4 & 85.3 & 512 & 51.5 & 72.7 \\
\midrule
\multicolumn{3}{c}{\textit{Bad Threshold}} & \multicolumn{3}{c}{\textit{Trajectory Selection}} \\
$\underline{r} = -0.3$ & 61.8 & 86.2 & $|r| > \bar{r}$ & \textbf{68.1} & \textbf{94.0} \\
$\underline{r} = 0$ & \textbf{68.1} & \textbf{94.0} & $|\rho_{\text{max}}| > \bar{\rho}$ & 68.1 & \textbf{94.0} \\
$\underline{r} = 0.3$ & 64.1 & 88.5 & $a \neq \{\}$ & 51.5 & 72.7 \\
\midrule
\multicolumn{3}{c}{\textit{SFT Frequency}} & \multicolumn{3}{c}{\textit{Reward Shaping}} \\
No SFT & 58.9 & 77.5 & $K=0.2$ & 68.0 & 92.6 \\
One-time & 64.3 & 80.4 & $K=0.3$ & 64.3 & 87.5 \\
Three-time & \textbf{68.1} & \textbf{94.0} & $K=0.4$ & \textbf{68.2} & \textbf{94.0} \\
\bottomrule
\end{tabular}
}
\end{table}

Furthermore, we conducted an in-depth analysis of key hyperparameters 
 within \(f_{\text{LLM}}\) and  \(g_{\text{LLM}}\) in Table \ref{tab:llm_ablation}.

For \(f_{\text{LLM}}\), we analyze three key aspects: (1) we test \(f_{\text{LLM}}\) query intervals of 128, 256, and 512 training steps with \(g_{\text{LLM}}\) query interval fixed at 32 to assess the impact of  \(f_{\text{LLM}}\)'s activation frequency on convergence and efficiency; (2) we explore bad-case thresholds \{-0.3, 0, 0.3\} to regulate LLM-refined sample volume; and (3) we evaluating ``No SFT'', ``One-time SFT'' at RL’s 2000th epoch, and ``three-time SFT'' per 100 refined samples to quantify the fine-tuning effects on performance.

For \(g_{\text{LLM}}\), we also analyze three key aspects: (1) we test \(g_{\text{LLM}}\) query intervals of 32, 128, and 512 training steps to assess the impact of \(g_{\text{LLM}}\)'s \textit{activation frequency}; (2) given the impracticality of using complete trajectories for LLM-based reward shaping, we explore three key trajectories selection criteria to reduce \(g_{\text{LLM}}\) input tokens: \textbf{reward-based} selection filter trajectories by reward absolute values to capture high- and low-reward cases, \textbf{state-based} selection identify trajectories where maximum line flow change exceeds a threshold \(|\rho_{\text{max}}| > \bar{\rho}\), and \textbf{action-based} selection designate trajectories with action changes \(a \neq \{\}\) as key trajectories; and (3) to prevent policy oscillation in Q-value estimation, we test shaping parameters \(K = \{0.2, 0.3, 0.4\}\) with \(g_{\text{LLM}}\) fixed at a 32-step query interval.

The experimental results highlight several critical insights:

\textit{(1) Higher activation frequency leads to faster initial RL learning}. A 256-step interval of \(f_{\text{LLM}}\) and 32-step interval of \(g_{\text{LLM}}\) shows significant improvements over 512 steps. Interestingly, as the number of training scenarios increased from 288 to 576 cases, the performance gap between different frequencies narrowed, suggesting that we can reduce activation frequency for environments with sufficiently diverse scenarios while maintaining effectiveness.

\textit{(2) Extreme thresholds for filtering bad cases are suboptimal.} $\underline{r}=0.3$ introduced 510 samples, $\underline{r}=0$ included 275 samples, and $\underline{r}=-0.3$ allowed only 83 samples for refinement. In our experiments, constrained by the reasoning capability of the base model, introducing too many samples results in minimal refinement benefits while increasing \(f_{\text{LLM}}\) reference time by 46\%. Conversely, too few samples led to slower convergence and approximately 6\% performance degradation compared to the standard settings. 

\textit{(3) Both reward-based and state-based key trajectory selection criteria in \(g_{\text{LLM}}\) improve performance}, while action-based selection shows significantly limited performance. This suggests that LLMs are more effective at extracting meaningful patterns from explicit information, such as performance indicators or state changes, compared to abstract topological changes.

\textit{(4) SFT frequency directly impacts ACE performance.} No SFT showed limited improvement after about 2000 epochs, with the survival rate stabilizing at 77.5\%. Applying one-time SFT led to immediate improvement, reaching 78.5\% in the first post-SFT evaluation step. Moreover, multiple SFT iterations demonstrated further enhancement over single SFT, ultimately achieving a survival rate of 84.8\%.

\subsection{Computational Overheads}

We conducted experiments on the NeurIPS 2020 competition environment to evaluate the computational efficiency of ACE. For expert-guided RL, we trained for 100K timesteps with a total duration of 6h 4m 14s. Table~\ref{tab:computational_overhead} presents the detailed computational costs and memory requirements of ACE using Qwen2-7B with SFT.

\begin{table}[htbp]
\centering
\caption{Computational overhead breakdown for ACE training}
\label{tab:computational_overhead}
\begin{tabular}{lccc}
\toprule
\textbf{Module} & \textbf{Counts} & \textbf{Samples} & \textbf{Time} \\
\midrule
ACE-RL & - & $\sim$40K & 3h 4m 41s \\
ACE-LLM Inference & 508 & 264 & 1h 48m 0s \\
ACE-LLM Sampling & 4981 & 32 & 59m 12s \\
ACE-LLM Training & 2 & 200 & 26m 10s \\
\midrule
\textbf{ACE-Total} & - & $\sim$40K & \textbf{6h 18m 3s} \\
\bottomrule
\end{tabular}
\end{table}

ACE's additional computation primarily stems from three LLM modules: selective inference, sampling, and training. With $f_{\text{LLM}}$ and $g_{\text{LLM}}$ query intervals set to 256 and 32 respectively, only 264 samples and 508 inferences were required during ACE training. The SFT module is executed only twice during the entire co-training process. Unlike traditional LLM-RL collaboration that requires continuous LLM interaction, ACE maintains learning efficiency by sampling from the constructed LLM buffer even when $f_{\text{LLM}}$ and $g_{\text{LLM}}$ are inactive, significantly improving sample efficiency.

\section{Discussion}
\subsection{Why LLMs Can Guide RL but Struggle in Sequential Industrial Decision-Making?}

\begin{figure}[t]
\begin{center}
\centerline{\includegraphics[width=\columnwidth]{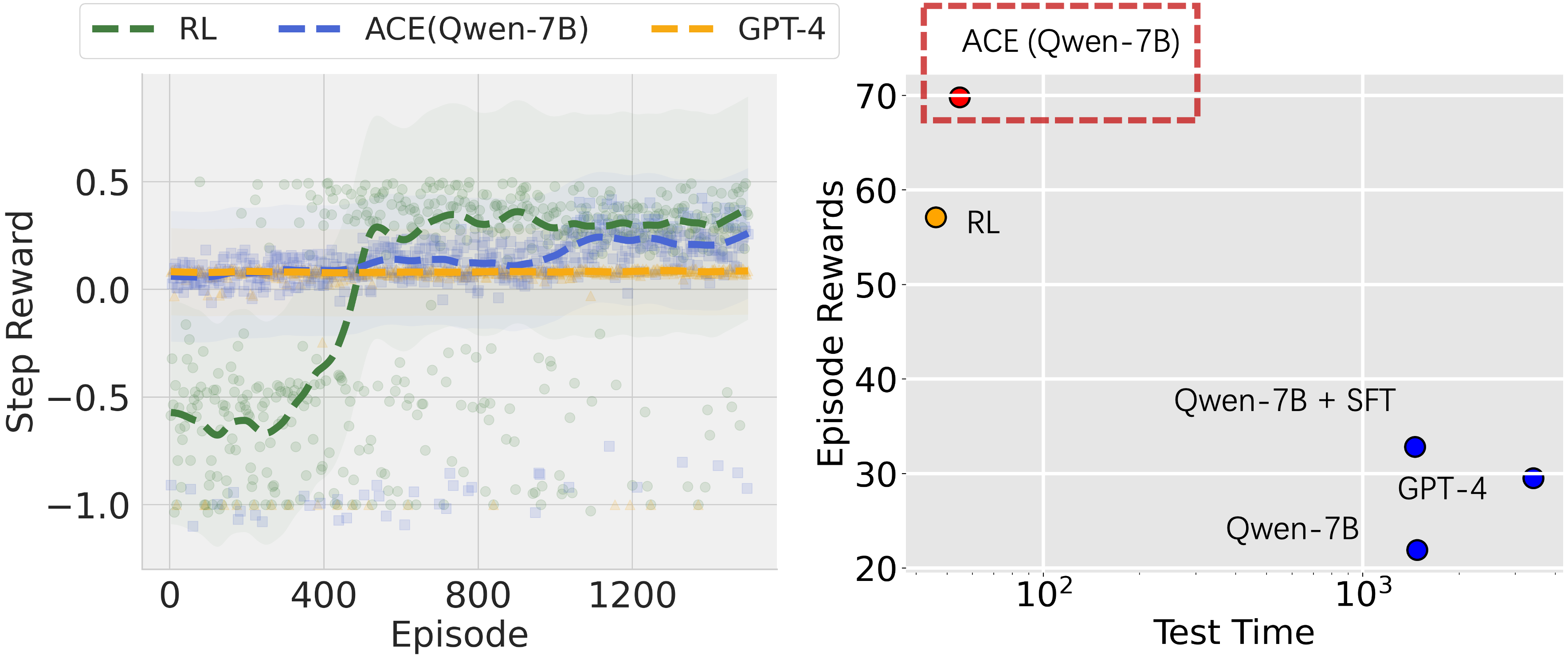}}
\caption{Left: Step-wise reward analysis in refining RL  trajectories. Right: Comparison between episode performance and test time for different RL and LLM configurations on WCCI 2020.}
\label{fig:Discussion}
\end{center}
\vskip -0.2in
\end{figure}

As illustrated in Figure \ref{fig:Discussion} (a), LLMs exhibit strong single-step reasoning and error-correction capabilities, especially in RL's early training stages. However, as shown in Figure \ref{fig:Discussion} (b), LLMs struggle to independently complete all decisions in an entire episode (approximately 800-2000 consecutive decision steps). This is because industrial data often involves control delays and high temporal coupling, while LLMs have difficulty modeling long-horizon dependencies due to their token-based memory bottleneck. Extending context windows to retain trajectory history increases inference latency quadratically, violating real-time constraints.

Furthermore, utilizing the mixed buffer constructed from RL and LLM can significantly enhance LLM's decision-making and correction capabilities. As shown in Figure \ref{fig:Discussion} (a), where ACE (Qwen2-7B+SFT)'s refined step reward gradually increases after SFT while GPT4's refined reward remains constant. Additionally, with $\mathcal{D}_{\text{mix}}$ fine-tuning, Qwen2-7B+SFT's independent decision-making performance improves by 10.9\% compared to the version without SFT, surpassing that of GPT4.

\subsection{Why using LLMs for Trajectories Refinements is Better Than Policy Regularization?}

LLM4Teach uses KL-divergence constraints to enforce static alignment between LLM priors and RL policies. However, as demonstrated in Figure~\ref{fig:Discussion}, pure LLM policies are not well-suited for sequential decision-making in industrial scenarios. This policy-level alignment inevitably leads to over-conservative decision-making, as it forces RL policies to match LLM behaviors that are suboptimal for long-horizon control. In contrast, the ACE framework, by reprocessing trajectories offline, performs targeted reasoning on key cases to identify critical decision points. The test performance indicates that this selective refinement approach is more suitable for large-scale industrial scenarios.

\section{Conclusion}
This paper introduces ACE, a framework that synergizes LLMs and RL for large-scale industrial control through a ``\textit{Think Twice, Act Once}'' mechanism. We highlight our contributions from two perspectives: (1) we develop a more suitable LLM-RL collaborative framework for industrial decision-making, leveraging offline LLM refinement for both action refine and reward reshaping; (2) we establish an effective workflow that combines RL interaction and LLM refinement to generate high-quality mixed datasets, providing new insights for LLM applications in industrial tasks. Through extensive experiments on three large-scale power grid competitions, we demonstrate ACE's superior performance in terms of control effectiveness, decision-making efficiency, and sample efficiency. 


\section*{Impact Statement}
This paper presents work whose goal is to advance the field of Machine Learning. There are many potential societal consequences of our work, none of which we feel must be specifically highlighted here.
\section*{Acknowledgements}
This work was supported in part by the Natural Science Foundation of Zhejiang Province under Grant LZ23F030009, and the National Natural Science Foundation of China under Grants 52161135201. 


\bibliography{main}
\bibliographystyle{icml2025}

\newpage
\appendix
\onecolumn
\section{Appendix / Experiment Setting}
\subsection{Environment description}
\label{app:env}
\begin{figure}[ht]
\vskip 0.2in
\begin{center}
\centerline{\includegraphics[width=\columnwidth]{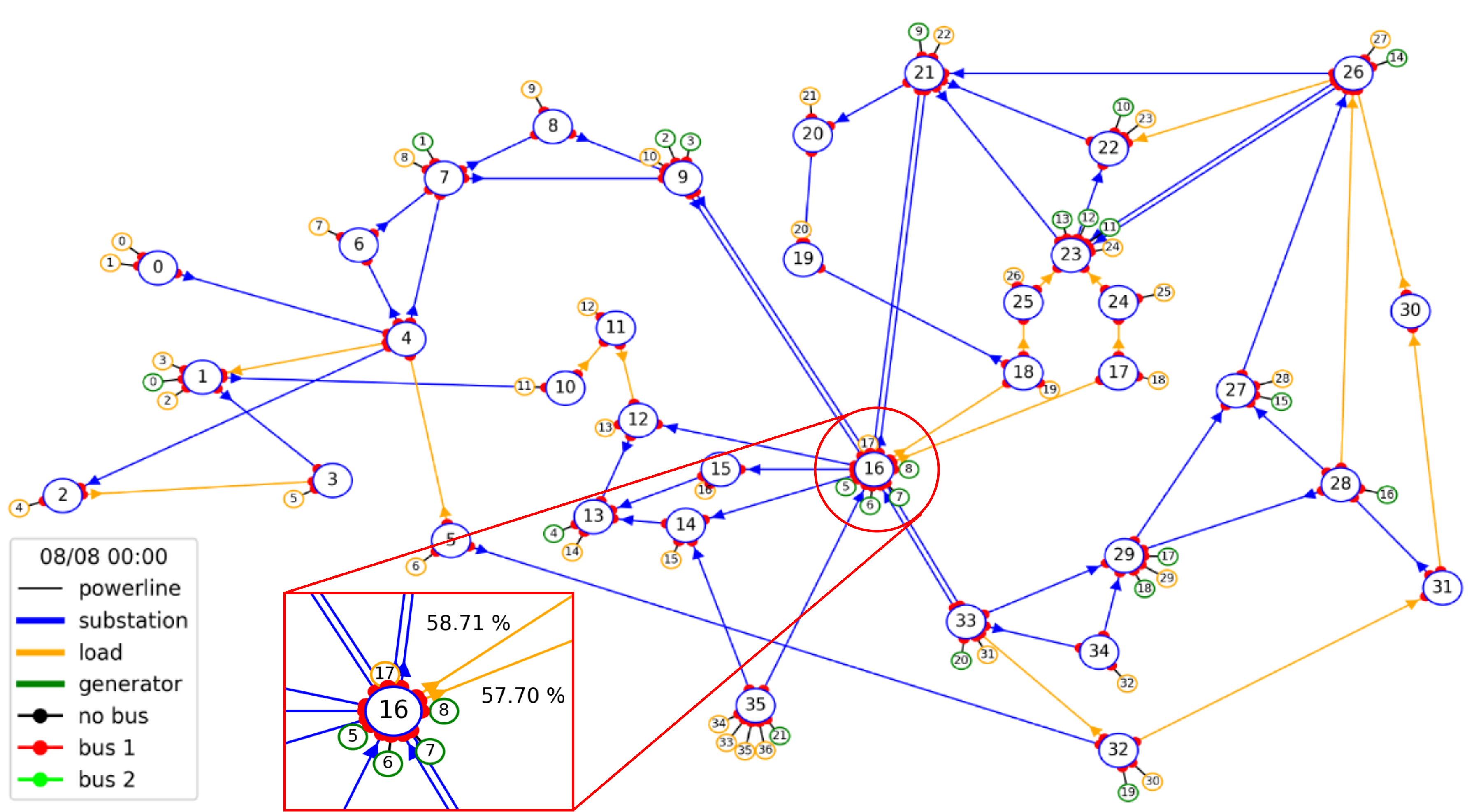}}
\caption{The test environment of L2RPN WCCI 2020 and NeurIPS 2020 challenge, consisting of 36 substations (blue circles), 59 transmission lines (blue lines), 22 generators (green marks) and several loads (yellow marks). Red boxes highlight critical load areas with percentage indicators showing the operational status of power lines. }
\label{fig:Environment}
\end{center}
\vskip -0.2in
\end{figure}

We evaluate the ACE framework on the Grid2Op platform \cite{donnot2020grid2op}, which provides an industry-standard power grid simulation environment, as shown in Figure~\ref{fig:Environment}. 

Below, we describe the key components of the MDP formulation for ACE setting.

\paragraph{State Space $\mathcal{S}$} 
The power grid state is characterized by six carefully selected features from the 12 available observation variables.
The detailed state is shown in Table~\ref{tab:observation}.
These features are selected based on the prior work \cite{yoon2021winning} to provide a compact yet comprehensive representation of the grid's operational state.

\begin{table}[ht]
\centering
\label{tab:observation}
\caption{Details of Grid2Op environment observation space}
\resizebox{1\textwidth}{!}{ 
\begin{tabular}{llll}
\toprule
Name & Type & Size & Description \\
\midrule
Date & int & 6 & The current year, month, day, hour of day, minute of hour, day of week. \\
Active power & float & $N_{\text{gen}} + N_{\text{load}} + 2 \times N_{\text{line}}$ & Active power magnitude for all elements. \\
Reactive power & float & $N_{\text{gen}} + N_{\text{load}} + 2 \times N_{\text{line}}$ & Reactive power magnitude for all elements. \\
Voltage & float & $N_{\text{gen}} + N_{\text{load}} + 2 \times N_{\text{line}}$ & Voltage magnitude at each bus. \\
Rho & float & $N_{\text{line}}$ & The capacity ratio between current flow and thermal limit for each line. \\
Topology & int & $N_{\text{gen}} + N_{\text{load}} + 2 \times N_{\text{line}}$ & Bus connectivity status for each element in its corresponding substation. \\
Line status & bool & $N_{\text{line}}$ & Binary indicator for line connection status. \\
Time step overflow & int & $N_{\text{line}}$ & Duration counter for line overflow events. \\
Time before cooldown line & int & $N_{\text{line}}$ & Remaining cooldown time before lines become operable again. \\
Time before cooldown sub & int & $N_{\text{sub}}$ & Remaining cooldown time before substations become operable again. \\
Time next maintenance & int & $N_{\text{line}}$ & Time until next scheduled maintenance for each line. \\
Duration next maintenance & int & $N_{\text{line}}$ & Expected duration of next maintenance period. \\
\bottomrule
\end{tabular}
}
\end{table}

\paragraph{Action Space $\mathcal{A}$}
The control actions available to the agent fall into two categories:
\begin{itemize}
   \item \textit{Bus assignment}: modifying element connections within substations
   \item \textit{Line switch}: controlling the connection status of transmission lines
\end{itemize}
Given $N_{\text{line}}$ transmission lines, $N_{\text{sub}}$ substations, and $\text{Sub}(i)$ elements in the $i$-th substation, the action space cardinality is:
\begin{equation}
   |{\mathcal A}| = N_{\text{line}} + 2^2 \times N_{\text{line}} + \sum_{i=0}^{N_{\text{sub}}} 2^{\text{Sub}(i)}
\end{equation}

For WCCI 2020 and NeurIPS 2020 challenges, the cardinality exceeds 60,000 possible actions, while for the WCCI 2022 challenge with the IEEE-118 bus system, it grows further to over 70,000 actions \cite{yoon2021winning,dorfer2022power}.

\paragraph{Reward Function $\mathcal{R}$}
Inspired by the setup of the WCCI 2020 winning solution \cite{yoon2021winning,dorfer2022power}, we design a reward mechanism that integrates dual objectives of operational efficiency and system security:

\begin{equation}
   r_t = \begin{cases}
   \frac{\text{load}_t}{\text{prod}_t} & \text{if not} \ d_t \\
   -\lambda_{\text{fail}} & \text{else}
   \end{cases}
\end{equation}
where $\frac{\text{load}_t}{\text{prod}_t}$ represents the instantaneous grid efficiency at time step $t$. However, if demand exceeds production ($\frac{\text{load}_t}{\text{prod}_t} > 1$), it triggers an immediate episode termination with a substantial penalty $\lambda_{\text{fail}}$.

\subsection{Hyper-parameters}
\label{app:parameters}
We show the hyperparameters used in our experiments, as detailed in Table \ref{tab:hyper}.

\begin{table}[ht]
\centering
\label{tab:hyper}
\caption{RL-LLM Collaborative Decision Parameters}
\begin{tabular}{llll}
\toprule
\textbf{Parameter} & \textbf{Value} & \textbf{Parameter} & \textbf{Value} \\
\midrule
\multicolumn{4}{l}{\textit{RL Parameters}} \\
\midrule
Replay buffer size & 50,000 & Training episodes & 100,000 \\
State embedding dim & 128 & Attention heads & 8 \\
History window & 6 & Batch size & 64 \\
Discount ($\gamma$) & 0.995 & Learning rate & 5e-5 \\
Target update interval & 2 & Soft update ($\tau$) & 1e-3 \\
Max episode length & 864 / 2016 & Samples drawn ratio $\beta$ & 0.5 \\

\midrule
\multicolumn{4}{l}{\textit{LLM Parameters}} \\ 
\midrule
LLM guidance buffer & 256 & LLM response budget & 512 / 1024 \\
Actor query interval & 256 & Critic query interval & 32 \\
SFT buffer size & 10,00 & SFT batch size & 8 \\
Adjustment scale $K$ & 0.2 / 0.5 & Reward threshold ($\bar{r} \& \underline{r}$)& 0.1 \& 0.5 \\
Max Token of $f_{\text{LLM}}$ & 5120 & Max Token of $g_{\text{LLM}}$ & 4096 \\
\bottomrule
\end{tabular}
\end{table}
\section{Appendix / More Detail of ACE}
\label{app:ACE}
\subsection{Agent Prompt Detail}

\begin{examplebox}{Input State-Action Parsing for $f_{\text{LLM}}$}
You are an expert power grid operator. Now, Let's analyze the current situation (2012-1-15 00:30) step by step:

\textbf{Grid Overview}:

    Total elements: 177
    
    Operable substations: [16, 23, 26, 21, 9, 29, 33, 35, 1, 4, 7]
    
    Total lines: 59

\textbf{Overload Lines:}

    - Line id 21 (Usage: 96.30\%) connects Substation 15 and Substation 16
    
    - Line id 23 (Usage: 98.43\%) connects Substation 16 and Substation 18
    
    - Line id 24 (Usage: 118.57\%) connects Substation 18 and Substation 19
    
    - Line id 25 (Usage: 163.69\%) connects Substation 19 and Substation 20
    
    - Line id 26 (Usage: 95.56\%) connects Substation 20 and Substation 21
    
    - Line id 38 (Usage: 185.42\%) connects Substation 23 and Substation 26
    
    - Line id 39 (Usage: 101.19\%) connects Substation 22 and Substation 26
    
    No disconnected line!

\textbf{Crucial Substations:}

    Substation id 16 current topology:
    
    \quad - Lines connected in Bus 0: {17 (Usage: 32.23\%), 18 (Usage: 54.52\%), 19 (Usage: 0.00\%), 20 (Usage: 42.02\%), 21 (Usage: 96.30\%), 22 (Usage: 67.58\%), 23 (Usage: 98.43\%), 27 (Usage: 34.86\%), 28 (Usage: 34.99\%), 48 (Usage: 23.52\%), 49 (Usage: 23.52\%), 54 (Usage: 40.54\%)}
    
    \quad - Lines connected in Bus 1: \{\}
    
    \quad - Lines disconnected: \{\}
    
    Substation id 21 current topology:
    
    \quad - Lines connected in Bus 0: {26 (Usage: 95.56\%), 27 (Usage: 34.86\%), 28 (Usage: 34.99\%), 29 (Usage: 116.08\%), 30 (Usage: 44.63\%), 36 (Usage: 71.57\%)}
    
    \quad - Lines connected in Bus 1: \{\}
    
    \quad - Lines disconnected: \{\}
    
    Substation id 23 current topology:
    
    \quad - Lines connected in Bus 0: {30 (Usage: 44.63\%), 31 (Usage: 175.90\%), 32 (Usage: 56.19\%), 34 (Usage: 60.74\%), 37 (Usage: 53.60\%), 38 (Usage: 185.42\%)} 
    
    \quad - Lines connected in Bus 1: \{\}
    
    \quad - Lines disconnected: \{\}
    
    Substation id 26 current topology: 
    
    \quad - Lines connected in Bus 0: {36 (Usage: 71.57\%), 37 (Usage: 53.60\%), 38 (Usage: 185.42\%), 40 (Usage: 119.37\%), 41 (Usage: 194.56\%), 56 (Usage: 29.90\%)}
    
    \quad - Lines connected in Bus 1: {39 (Usage: 101.19\%)}
    
    \quad - Lines disconnected: \{\}

\textbf{Bad Line Change Examples: }

    Please AVOID the Line change: \{\} as it is a BAD action because it results in a reward of -1.0.

\textbf{Operational Constraints}

    No lines in cooldown
\end{examplebox}

\begin{examplebox}{Task Description Prefix of $f_{\text{LLM}}$}
\textbf{Important Notes:}

1. Limit your changes no more than 5 lines id.

2. Consider adjusting the topology of their shared (connected) lines. This indirect approach may help redistribute the load and reduce stress on overloaded lines.

3. Reason from the example line changes, and avoid outputting the BAD line change.

\textbf{Response Format:}

Please analyze the situation and provide your response in the following format:

1. Analysis of critical issues.

2. Reason and analysis why the provided line change examples are BAD if provided.

3. Propose your response to target line changes (Use ONLY values 0 or 1 in bus\_id.)] proposed line changes: \{line\_id: new\_bus\_id, line\_id: new\_bus\_id \}

\textbf{Remember:}

1. Use exactly the format shown above. Do not add any bold formatting, asterisks, or other special characters.

3. For proposed line changes, only include the chosen lines and the target topology.

4. Use ONLY values 0 or 1 in new\_bus\_id.

5. Consider line cooldown constraints.

\end{examplebox}

\begin{examplebox}{Input State-Action Parsing for $g_{\text{LLM}}$}

\textbf{Episode Overview:}

    - Total steps: 14
    
    - Initial time step: 2012-4-23-6-55
    
    - Final time step: 2012-8-23-5-15
    
    - Cumulative reward: 34.46

\textbf{Key Timestep Analysis:}

    - Time step 2012-4-23-6-55:
    
         \quad Action: \{73: 1\} Reward: -0.48
         
         \quad Highest line usage: 90.68\% (Line 39) Overloaded lines: 0
         
         \quad Key reason: First/Last step, Significant change in highest usage
         
    - Time step 2012-4-21-5-55:
    
         \quad Action: \{58: 0\} Reward: 0.90
         
         \quad Highest line usage: 424.50\% (Line 0) Overloaded lines: 2
         
         \quad Key reason: Significant reward change, Significant change in highest usage, Change in number of overloaded lines, Topology change
         
    - Time step 2012-4-21-5-20:
    
         \quad Action: \{6: 0\} Reward: 0.58
         
         \quad Highest line usage: 128.16\% (Line 13) Overloaded lines: 2
         
         \quad Key reason: Significant change in highest usage, Topology change
         
    - Time step 2012-4-21-0-30:
         \quad Action: \{15: 1, 17: 0, 20: 1\} Reward: 4.58
         
         \quad Highest line usage: 90.55\% (Line 15) Overloaded lines: 0
         
         \quad Key reason: Significant reward change, Significant change in highest usage, Change in number of overloaded lines, Topology change
         
    - Time step 2012-4-20-23-55:
    
         \quad Action: \{56: 0\} Reward: 0.56
         
         \quad Highest line usage: 71.97\% (Line 41) Overloaded lines: 0
         
         \quad Key reason: Significant reward change, Topology change

    - Time step 2012-4-21-6-50: 
    
         \quad Action: \{30: 1\} Reward: -0.73
         
         \quad Highest line usage: 125.50\% (Line 41) Overloaded lines: 1
         
         \quad Key reason: Significant reward change, Significant change in highest usage, Change in number of overloaded lines, Topology change
\end{examplebox}

\begin{examplebox}{Task Description Prefix of $g_{\text{LLM}}$}
\textbf{Remember:}

1. Key decision point indices are the time step indices in the episode (starting from 0), select up to 4 most important decision points
    
2. Reward adjustments can only be one of +0.4, +0.2, -0.2, -0.4
    
3. Both lists must be of the same length and correspond in order
    
\textbf{Response Format:}

Please analyze the below information and select decision points where the reward estimation might be erroneous. Provide your analysis results in the following format:
    
Key Decision Point Indices: [X, Y, A, B]
    
Reward Adjustments: [W, V, T, S]

1. Index X (Adjustment W): [Explain why this decision point is important and why this adjustment value was chosen]
    
2. Index Y (Adjustment V): [Explain why this decision point is important and why this adjustment value was chosen]
    
3. Index A (Adjustment T): [Explain why this decision point is important and why this adjustment value was chosen]
    
4. Index B (Adjustment S): [Explain why this decision point is important and why this adjustment value was chosen]
\end{examplebox}

While $f_{\text{LLM}}$ and $g_{\text{LLM}}$ serve different roles in ACE, their prompts share a structured design pattern consisting of two essential components:

\textbf{Task Description Prefix}: A comprehensive context header that provides necessary domain knowledge and operational guidelines. For $f_{\text{LLM}}$, this includes power grid topology rules and operational constraints, while for $g_{\text{LLM}}$, it focuses on reward assessment criteria and safety standards.
   
\textbf{Input State-Action Parsing}: Dynamic information extracted from current sample's key transitions. Specifically, for $f_{\text{LLM}}$: Detailed state-action parsing focusing on critical line loadings and substation configurations. For $g_{\text{LLM}}$: Trajectory-level analysis highlighting key decision points and their consequences.

\begin{figure}[t]
\vskip 0.2in
\begin{center}
\centerline{\includegraphics[width=\columnwidth]{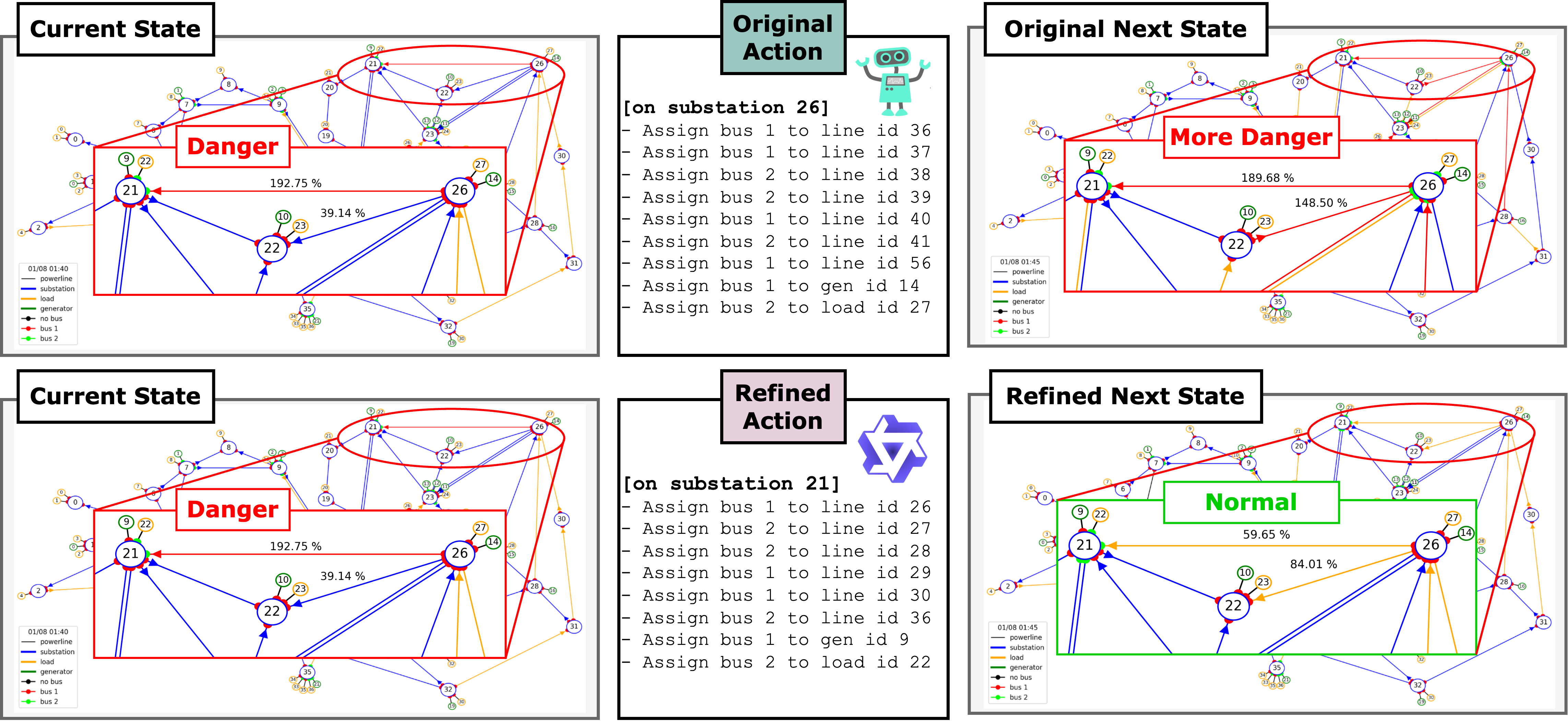}}
\caption{The demonstration of ACE action refinement in power grid operation. \textit{Upper row} shows the state transition under original RL action, where a dangerous state (left) becomes more severe (right) after applying the topology change on substation 26 (middle).\textit{ Lower row }demonstrates ACE's improvement, where the refined action on substation 21 successfully transitions the system from dangerous to normal operation.}
\label{fig:Demonstrate}
\end{center}
\vskip -0.2in
\end{figure}

\end{document}